\documentclass[sigconf]{acmart}
\usepackage{tabularx}
\usepackage{multirow}
\usepackage{multicol}
\usepackage{enumitem}
\usepackage{bbding}
\usepackage{bm}
\usepackage{makecell}
\usepackage{xcolor}
\usepackage{CJKutf8}
\renewcommand\footnotetextcopyrightpermission[1]{}

\usepackage[utf8]{inputenc}
\usepackage{array}

\AtBeginDocument{%
  \providecommand\BibTeX{{%
    \normalfont B\kern-0.5em{\scshape i\kern-0.25em b}\kern-0.8em\TeX}}}

\setcopyright{acmlicensed}
\copyrightyear{2018}
\acmYear{2018}
\acmDOI{XXXXXXX.XXXXXXX}

\acmConference[KDD2024]{ACM SIGKDD Conference on Knowledge Discovery and Data Mining}{August 25--29,
  2024}{Barcelona, Spain}
\acmISBN{978-1-4503-XXXX-X/18/06}

\begin{document}

\title{Towards Automatic Evaluation for LLMs' Clinical Capabilities: Metric, Data, and Algorithm}

\author{
    Lei Liu\textsuperscript{1,3,$\ddagger$} \,
    Xiaoyan Yang\textsuperscript{1,$\ddagger$} \, 
    Fangzhou Li\textsuperscript{2,$\ddagger$} \,
    Chenfei Chi\textsuperscript{2,$\ddagger$} \,
    Yue Shen\textsuperscript{1} \,
    Shiwei Lyu\textsuperscript{1} \,
    Ming Zhang\textsuperscript{2} \,\\
    Xiaowei Ma\textsuperscript{2} \,
    Xiangguo Lyu\textsuperscript{2} \,
    Liya Ma\textsuperscript{2} \,
    Zhiqiang Zhang\textsuperscript{1} \,
    Wei Xue\textsuperscript{2} \,
    Yiran Huang\textsuperscript{2,$\dagger$} \,
    Jinjie Gu\textsuperscript{1,$\dagger$}\\
    \textsuperscript{1}Ant Group \\
    \textsuperscript{2}Department of Urology, Shanghai Jiao Tong University School of Medicine Affiliated Renji Hospital\\
    \textsuperscript{3}The Chinese University of Hong Kong, Shenzhen\\
   \texttt{liulei1497@gmail.com, joyce.yxy@antgroup.com, huangyiran@renji.com}
}
\thanks{$\ddagger$ indicates equal contribution. $\dagger$ denotes the corresponding authors.}
\renewcommand{\shortauthors}{Lei Liu, Xiaoyan Yang et al.}
\begin{abstract}
Large language models (LLMs) are gaining increasing interests to improve clinical efficiency for medical diagnosis, owing to their unprecedented performance in modelling natural language. Ensuring the safe and reliable clinical applications, the evaluation of LLMs indeed becomes critical for better mitigating the potential risks, \textit{e.g.}, hallucinations. However, current evaluation methods heavily rely on labor-intensive human participation to achieve human-preferred judgements. To overcome this challenge, we propose an automatic evaluation paradigm tailored to assess the LLMs' capabilities in delivering clinical services, \textit{e.g.}, disease diagnosis and treatment. The evaluation paradigm contains three basic elements: metric, data, and algorithm. Specifically, inspired by professional clinical practice pathways, we formulate a LLM-specific clinical pathway (LCP) to define the clinical capabilities that a doctor agent should possess. Then, Standardized Patients (SPs) from the medical education are introduced as the guideline for collecting medical data for evaluation, which can well ensure the completeness of the evaluation procedure. Leveraging these steps, we develop a multi-agent framework to simulate the interactive environment between SPs and a doctor agent, which is equipped with a Retrieval-Augmented Evaluation (RAE) to determine whether the behaviors of a doctor agent are in accordance with LCP. The above paradigm can be extended to any similar clinical scenarios to automatically evaluate the LLMs' medical capabilities. Applying such paradigm, we construct an evaluation benchmark in the field of urology, including a LCP, a SPs dataset, and an automated RAE. Extensive experiments are conducted to demonstrate the effectiveness of the proposed approach, providing more insights for LLMs' safe and reliable deployments in clinical practice.

\end{abstract}



\keywords{Large Language Model, Evaluation Benchmark, Medical AI}

\maketitle

\section{Introduction}
With the capacity to generate human-like responses, LLMs could work as intelligent clinicians for medical consultations, largely enhancing the decision-making efficiency in clinical settings, such as medical education \cite{karabacak2023advent}, question-answering system \cite{singhal2023large,lee2023benefits}, and diagnostic dialogue \cite{stokel2023chatgpt}. However, LLMs are prone to a critical issue known as hallucination, wherein they fabricate erroneous judgments misaligned with clinical evidence \cite{zhang2023language}. This limitation poses a significant barrier to their deployment in the safety-critical clinical scenarios. Therefore, it is crucial to conduct a rigorous evaluation for LLMs’ medical capability and mitigate potential risks.

Towards the evaluation of LLMs’ medical capabilities, most existing approaches are task-oriented to measure the encoded specialized medical knowledge of LLMs. For this purpose, three tasks are generally utilized to indicate the quality of the medical knowledge within LLMs, \textit{i.e.}, medical information extraction (IE), medical question-and-answer (QA), and diagnosis dialogue. Specifically, medical IE is tasked with the extraction of pre-specified medical information from textual content, \textit{e.g.}, entity recognition. Medical QA and diagnostic dialogue focus on diagnostic accuracy via evaluating the word co-occurrence between LLM-generated answers and ground-truth references, for single-turn and multi-turn interactions, respectively. Furthermore, some recent studies \cite{shi2023llm,liao2023automatic} proposed to utilize patient simulator to achieve automatic evaluation based on the pre-defined medical skills. However, previous approaches may fail to evaluate LLMs' capabilities of being a clinician due to the following drawbacks: (1) The metrics primarily focus on the medical knowledge of LLMs, while ignoring the professional clinical practice pathways. (2) There lacks of the guidance for data collection, \textit{i.e.}, how to collect high-quality data for evaluation. (3) Most existing evaluation require human participation, which is time-consuming and labor-intensive.


\begin{figure*}[!t]
  \centering
  \includegraphics[width=0.85\textwidth]{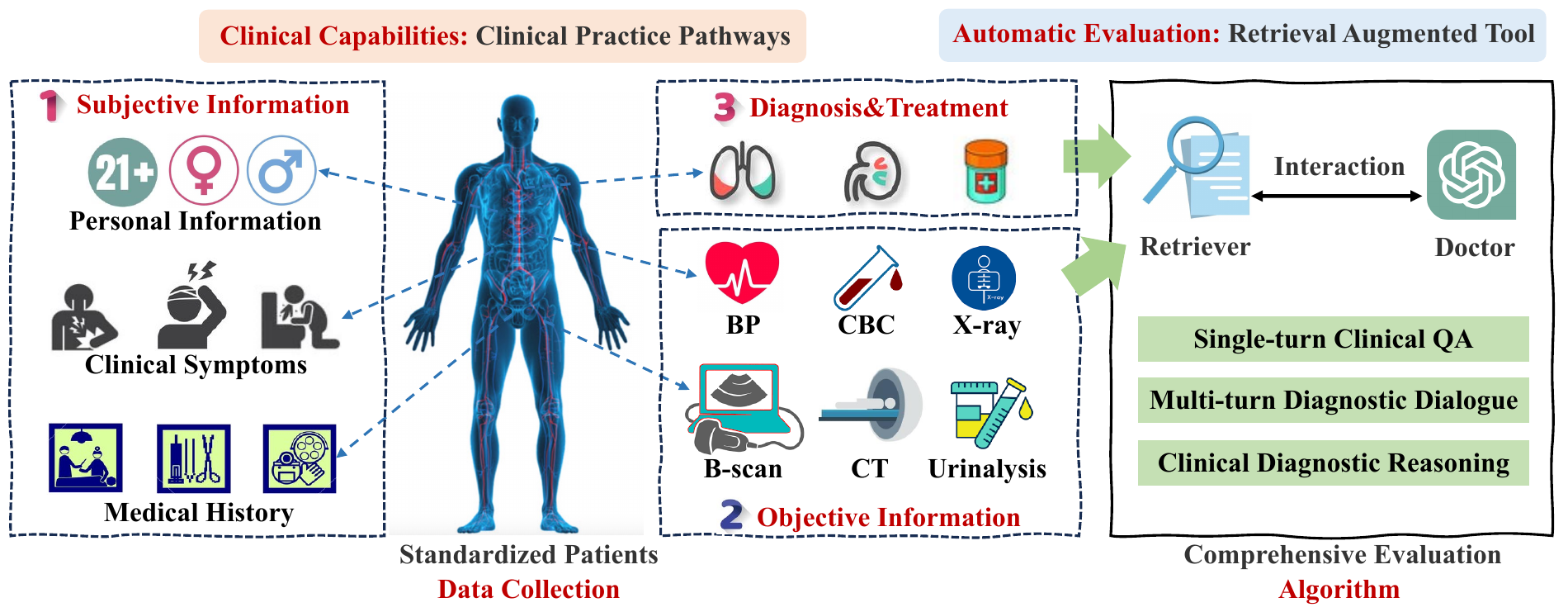}
  \caption{The proposed evaluation paradigm. {\color{red}\textbf{Metric:}} Clinical practice pathways are introduced from the medical education as the evidence of clinical capabilities. {\color{red}\textbf{Data:}} Standardized patients are taken as the template for data collection. {\color{red}\textbf{Algorithm:}} Retrieval-Augmented Evaluator can support a comprehensive and automated evaluation.}
  \label{fig:teaser}
\end{figure*}

From a high-level perspective, there indeed needs a comprehensive paradigm to establish an evaluation benchmark for LLMs' clinical capabilities, mainly including following aspects: \textbf{(1) \underline{Metrics}} define the clinical capabilities that are required for a doctor agent as a clinician. \textbf{(2) \underline{Data Collection}} indicates to how to collect high-quality evaluation data. \textbf{(3) \underline{Algorithm}} is to achieve automatic evaluations to avoid expensive expert labors. Previous studies primarily focused on one or more of these components, lacking an exploration of a comprehensive and macroscopic evaluation paradigm. To this end, we integrate the cutting-edge research outcomes from medical and AI fields to establish a paradigm to evaluate the clinical capability of LLMs, which contributes an explicit guidance for building a comprehensive evaluation benchmark from an interdisciplinary perspective.

Beginning from the realistic medical consultations in the hospitals, clinical practice pathways \cite{kinsman2010clinical,lawal2016clinical,de2018clinically} can provide the professional guidelines designed to deliver evidence-based healthcare with the essential steps. In detail, a basic clinical diagnosis pathway \cite{balogh2015improving} is defined as: At the first, a patient experiences a health problem. Then there is an iterative process of information gathering, information integration and interpretation, and determining a diagnosis. Performing a clinical history and interview, conducting a physical exam, and performing diagnostic testing are all ways of accumulating information that may be relevant to understanding a patient's health problem. Benefiting from the above medical knowledge, we formulate a general LLM-specific clinical pathway (LCP) to lay out a standardized diagnostic procedure, which guides LLMs to appropriately monitor the basic medical information of the patient. Specifically, a clinician committee was established to validate the correctness of LCP for real-world medical consultations, where 6 professional clinical practice capabilities are deemed necessary to provide reasonable diagnostic and treatment results, including \textbf{Information Completeness, Behavior Standardization, Guidance Rationality, Diagnostic Logicality, Treatment Logicality, and Clinical Applicability}.

With LCP in mind, standardized patients (SPs) from the medical education \cite{barrows1993overview} greatly inspire us to design the standards for data collection. SPs are the individuals specially trained to act as virtual patients for the assessment of medical examination skills of clinicians. When using real-world medical data, the diagnostic dialogue will be interrupted due to lacking of some test reports, because a patient is not required to perform all laboratory tests during real-world medical consultations. Naturally, a fundamental question arises: \textit{\textbf{How to maintain the completeness of SPs' medical records?}} The core solution is to apply SPs for the data collection with the clinician participation. First, large-scale medical records are collected from the hospital. Then, clinicians need to formulate SPs with virtual yet complete medical records by taking the statistic information of the collected records as reference, \textit{e.g.}, the age group of high incidence population. Thus, SPs can well ensure the completeness of the evaluation procedure.

The human participation in the existing evaluation approaches is labor-intensive, despite of the effectiveness of LCP and SPs. Motivated by the Retrieval Augmented Generation (RAG) technique, we develop a Retrieval Augmented Evaluation (RAE) to simulate the interactive environment with the doctor agent. In particular, RAE is a retrieval algorithm to evaluate whether a doctor agent possess the capabilities of a clinical physician in accordance with LCP, including: collecting medical information from patients, guiding patients for appropriate laboratory test, as well as providing accurate diagnostic and treatment results. 




Overall, in this work, we propose a comprehensive paradigm for the automatic evaluations of LLMs' clinical capabilities, which can be extended to any similar clinical scenarios. The proposed paradigm is characterized by the professional medical knowledge (\textit{i.e.}, clinical practice pathways and SPs) and AI techniques (\textit{i.e.}, RAG). Benefiting from the paradigm, we develop LCP as metrics, SPs as data, and RAE as an automatic algorithm to construct a comprehensive evaluation paradigm. The propose approach is successfully implemented to construct an evaluation benchmark in the field of urology, indicating the methodology effectiveness. In summary, the main contributions of our work are follows:
\begin{enumerate}
    \item We propose a comprehensive evaluation paradigm tailored to measure LLMs' clinical capabilities. To our knowledge, this is the first approach to guide how to establish an automatic evaluation framework for LLMs, from the viewpoints of metrics, data collection, and algorithm.

    \item  We formulate a general LLM-specific clinical pathway (LCP) to describe the clinical capabilities of LLMs during medical consultations (\textit{i.e.}, metric). Standardized patients are introduced to ensure the completeness of the evaluation procedure (\textit{i.e.}, data collection). Leveraging them, we develop a Retrieval-Augmented Evaluation (RAE) to simulate the interactive clinical environment for quantifying LLMs' clinical behaviors (\textit{i.e.}, algorithm).

     \item The propose paradigm is successfully implemented to construct an evaluation benchmark for medical specialty, which provides a comprehensive evaluation for LLMs' clinical capabilities in the field of urology. Extensive experiments demonstrate the effectiveness of the proposed LCP, SPs, and RAE, indicating the reliability of the paradigm.
\end{enumerate}

\section{Related Work}
Many studies attempts to apply LLMs as an intelligent doctor in the medical domain \cite{chen2023huatuogpt,yunxiang2023chatdoctor,li2023beginner}. To ensure safe and reliable clinical applications, it is essential to evaluate LLMs' capabilities \cite{singhal2023towards} for working as a clinician. For this purpose, lots of medical QA and multi-choice benchmarks are proposed to measure the medical knowledge of LLMs \cite{zakka2023almanac}, such as PubMedQA \cite{jin-etal-2019-pubmedqa} and MedMCQA \cite{pal2022medmcqa}. Among these methods, medical examination is generally utilized to evaluate the medical capabilities of LLM. For instance, the studies \cite{gilson2023does,kung2023performance} reported the performance of LLMs in medical examination assessment through the United States Medical Licensing Examination (USMLE). \cite{zhang2023huatuogpt,chen2023huatuogpt} performed a comprehensive comparisons for LLMs' medical capabilities through Chinese National Medical Licensing Examination. However, these methods omit the assessment of the professional clinical practice capabilities, failing to evaluate LLMs' capabilities of being a clinician. Besides, their evaluation procedures usually require human participation, which is time-consuming and labor-intensive.

To alleviate the above-mentioned difficulties, some prior studies proposed to utilize patient simulator to achieve automatic evaluation, which is based on the pre-defined medical skills. For example, \cite{shi2023llm} established a LLM-Mini-CEX approach to assess the diagnostic capabilities of LLMs. \cite{liao2023automatic} proposed an automatic evaluation benchmark for multi-turn consultations, where the data source of patients is from the medical QA datasets. However, there indeed lacks of an unified paradigm for the automatic evaluation from t he viewpoints of metric, data, and algorithm.

Compared to previous work, our proposed approach provides an unified paradigm with following innovative points: 1) The clinical capabilities of LLMs should be defined according to the
clinical practice pathways in the medical field. 2) The data completeness is important for collecting evaluation data, where SPs can support the doctor agent to complete the entire clinical diagnosis task. 3) Retrieval task is a feasible way to achieve an automatic evaluation algorithm by leveraging the pre-defined metrics and SPs' records.

\begin{figure}[!t]
  \centering
  \includegraphics[width=0.85\linewidth]{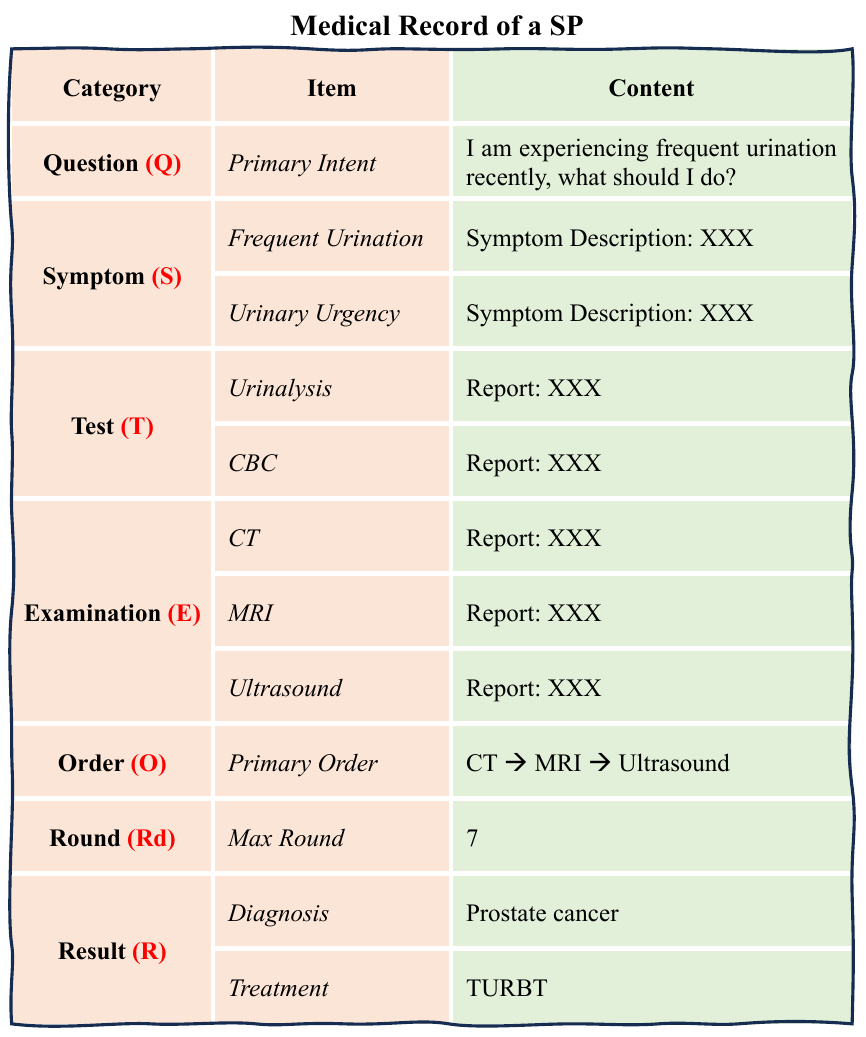}
  \caption{A simplified example of structural SPs' medical records. Some details are omitted due to the limited space, such as the report results (denoted by XXX). Category and item are for bi-level retrieval, respectively. The full-version example can be seen in Table \ref{tab:sps} of the Appendix.}
  \label{img:sps}
\end{figure}

\begin{table*}[!t]
\centering
  \caption{The implementation of the proposed paradigm. The capabilities are derived from the LCP, which induces the principles of data collection. RAE can achieve an automatic evaluation algorithm via the retrieval task.}
  \label{tab:detail}
  \begin{tabular}{l|l|l|p{5cm}}
    \toprule
    \textbf{Metric} & \textbf{SPs Data} & \textbf{Algorithm} & \textbf{Capability}  \\ \midrule
    Information Completeness & {\color{red}\textbf{(S)} \textbf{(T)} \textbf{(E)}}  &How much SPs' information is retrieved. &Enquire medical information of patients.\\\midrule
     Behavior Standardization &{\color{red}\textbf{(O)}} & Whether following the retrieved inquiry order.&Enquire information by a suitable order.\\\midrule
     Guidance Rationality  &{\color{red}\textbf{(T)} \textbf{(E)}}& How many reasonable tests/exams are retrieved.&Enquire reasonable test/exam reports. \\\midrule
    Diagnostic Logicality &{\color{red}\textbf{(R)}}&Generations \textit{\textbf{VS.}} Retrieved ground-truth.&Reason out the correct diagnosis results.\\\midrule
    Treatment Logicality &{\color{red}\textbf{(R)}}&Generations \textit{\textbf{VS.}} Retrieved ground-truth.&Reason out the correct treatment plans.\\\midrule
    Clinical Applicability      &{\color{red}\textbf{(Rd)}}& Agent's round \textit{\textbf{VS.}} Retrieved clinician's round & Finishing tasks within reasonable consultation rounds.\\
  \bottomrule
  \end{tabular}
  \label{tab:system}
\end{table*}

\begin{figure*}[!t]
  \centering
  \includegraphics[width=1\linewidth]{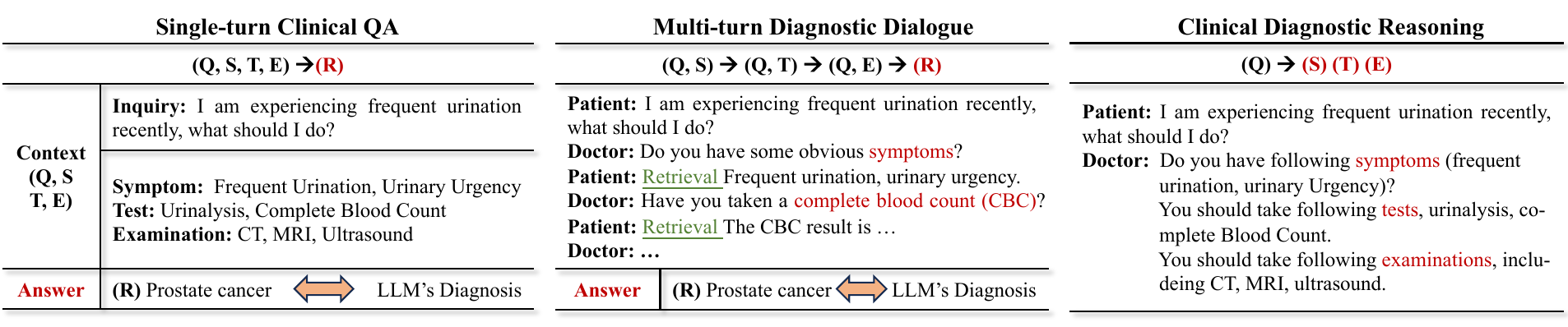}
  \caption{Example of different medical tasks. The retrieval task (\textbf{green font}) is used to construct data format, which can be further exploited for automatic evaluations (\textbf{red font}). The data source for the retrieval task is SPs data, as shown in Figure \ref{img:sps}.}
  \label{fig:tasks}
\end{figure*}

\section{The proposed Paradigm}
In this section, we will first introduce the evaluation paradigm, including LCP, SPs, and RAE. Then we will introduce the proposed benchmark for evaluating LLMs in the field of urology.

\subsection{LLM-specific Clinical Pathway}

\subsubsection{Clinical Practice Pathways}
Clinical practice pathways \cite{kinsman2010clinical,lawal2016clinical,de2018clinically} provide the professional guidelines to deliver evidence-based healthcare with the essential steps, mainly including the follows.

\textit{\textbf{Self-report Collection.}}
Self-report refers to the patients' personal information about their own health condition, daily lifestyle, and any other healthy aspects, which might not be observable or measurable through clinical laboratory tests.

\textit{\textbf{Subjective Information Collection.}}
Self-report provides a coarse description of patient's health conditions rather than a comprehensive statement, since they are not aware of the importance of different symptoms. Doctors should be able to guide patients to recall some necessary information about their symptoms, feelings, perception of their health, and personal experiences of illness.

\textit{\textbf{Objective Information Collection.}}
In general, due to the limitations of self-reporting, \textit{e.g.}, information bias caused by subjective nature of the patients, a professional doctor needs to corroborate patient self-reports via clinical assessments and diagnostic tests, which is referred as objective information. Objective information is vital because it often provides tangible evidence that helps confirm a diagnosis or monitor the effectiveness of treatments. It is less influenced by personal interpretation or bias and is therefore considered a more reliable source of data regarding a patient's health status. \textbf{However, existing evaluation approaches may fail to measure such capability of LLMs working as virtual doctors.} During a medical consultation, objective information usually includes:
\begin{itemize}[leftmargin=*]
    \item Vital Signs: Temperature, blood pressure, heart rate, respiratory rate, and oxygen saturation.
    \item Physical Examinations: Observations from a physical exam, such as inspection, palpation, percussion, and auscultation.
    \item Laboratory Tests: Blood tests, urine tests, and other bodily fluids.
    \item Imaging Studies: X-rays, CT, MRI, ultrasounds, and other tests.
\end{itemize}

\textit{\textbf{Final Diagnosis and Treatment.}}
Both subjective and objective information gathered provides a context for understanding the patient's health issue. The doctors can combine their clinical experiences with medical knowledge to assess the patient's condition, make a diagnosis, and develop a treatment plan.  Previous evaluation approaches mainly focus on the accuracy measurements of diagnosis and treatment results.

\subsubsection{Clinical Capability}
A clinician committee is established to validate the correctness of LCP for real-world medical consultations. 6 professional clinical capabilities are deemed necessary for reasonable diagnostic and treatment results, including: \textit{information completeness, behavior standardization, guidance rationality, diagnostic logicality, treatment logicality, and clinical applicability}. The details of these capabilities are described in Table \ref{tab:detail}. These capabilities can be evaluated via the proposed RAE, which will be introduced in Section \ref{sec:RAE}.


\begin{figure*}[!t]
  \centering
  \includegraphics[width=0.95\linewidth]{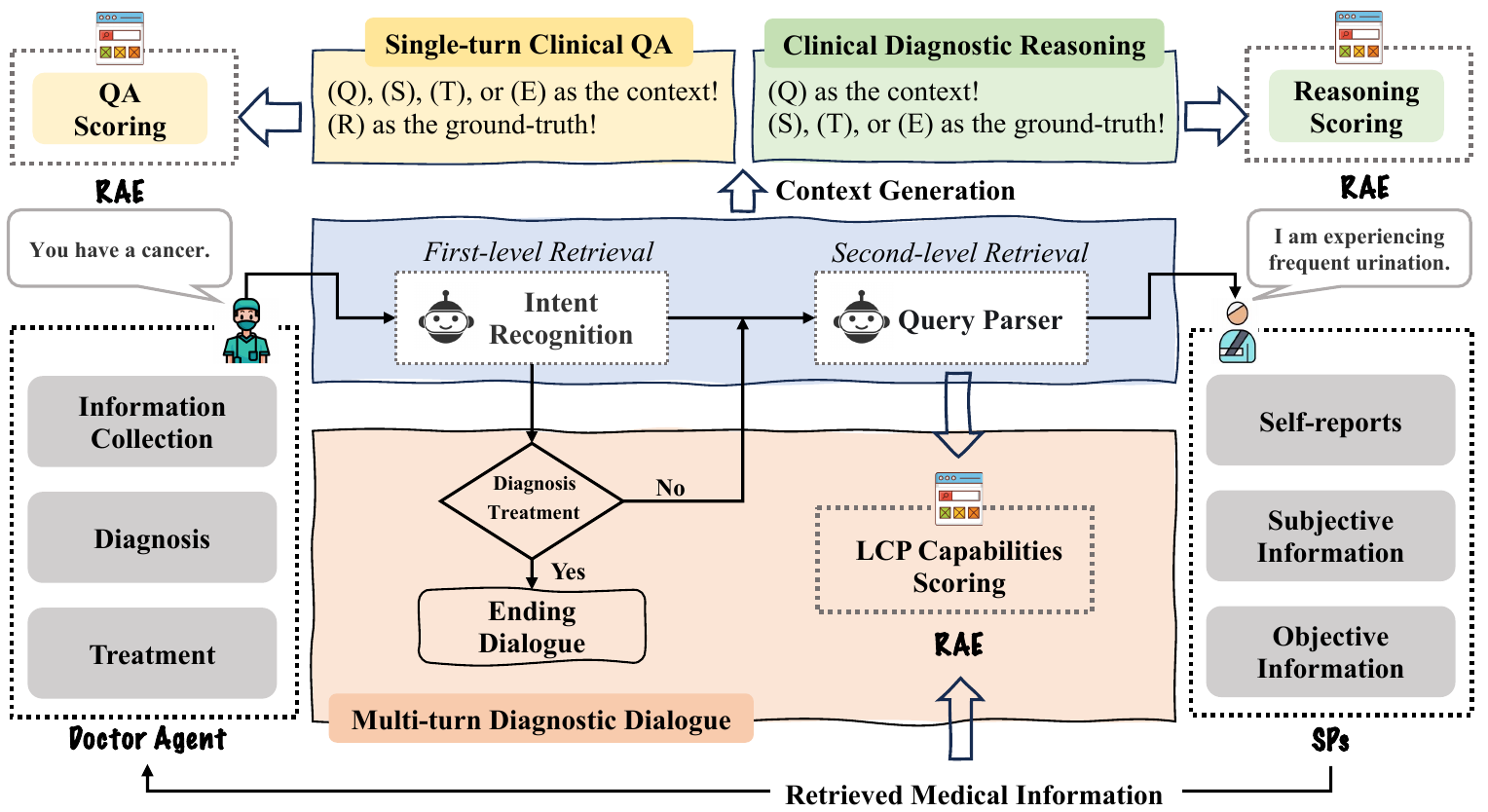}
  \caption{Overview of the multi-agent framework. Intent recognition aims to understand the doctor agent's query for terminating conversation. Query parser can map the doctor agent's query to bi-level structure. The multi-agent framework can achieve context generation for the clinical QA and reasoning tasks, as well as environment simulation for the diagnostic dialogue tasks. Besides, RAE can automatically evaluate the doctor agent's clinical capabilities.}
  \label{fig:multi-agent}
\end{figure*}


\paragraph{\textbf{Discussion}} Since different specialties may have different clinical pathways or standards, it is recommend to conduct some minor adjustments for LCP to accommodate various diseases.

\subsection{Standardized Patients}
\label{sec:sps}
Inspired by LCP and OSCEs, we introduce standardized patients (SPs) \cite{barrows1993overview} from medical education as references to collect evaluation data, where SPs are individuals who are trained to consistently and accurately simulate the symptoms, behaviors, and emotions of real patients in a reproducible way. One of the advantages of SPs is to well ensure the completeness of the evaluation procedure.

\subsubsection{Completeness} The main challenge of modeling SPs is to maintain the completeness of SPs' medical records, where realistic medical records cannot be directly applied as SPs' data. The main reason is that incomplete medical information of a medical record may lead to a failed evaluation. To be more clear, during an evaluation, SPs should perform a diagnostic dialogue with a doctor LLM. If SPs cannot handle doctor's normal inquiries about their personal medical information, the diagnostic dialogue would be interrupted, which leads to an unfinished evaluation process. For example, the doctor agent recommends that the patient should have a brain CT scan and the diagnostic dialogue would be terminated without final diagnostic results. Here, we provide a definition of the completeness of SPs' medical records as follows. 
\begin{definition}
    (Completeness) The medical record of a SP is termed as completeness if and only if this record contains all medical aspects related to the final diagnosis results, which may include but not limited to self-reports, subjective/objective information, and final diagnosis and treatments.
\end{definition}

\subsubsection{Bi-level Data Structure} 
According to the definition, the main difficulty of modeling SPs is to maintain the completeness of medical records. Therefore, we adopt to collect and manage structural SPs information for enforcing data consistency and accuracy.

As shown in in Figure \ref{img:sps}, we formulate a bi-level data structure to collect SPs information for evaluations: (1) First-level structure is for the coarse-grained categories of SPs' information, such as symptom, test, and examination. (2) Second-level structure contains the fine-grained items of first-level categories, such as CT and MRI. The details can refer to the 1-st and 2-nd categories in Table \ref{tab:sps} of the Appendix. Using such constraints can reduce the likelihood of errors and anomalies, which may be induced by human participation. Besides, SPs are stored in well-defined structure, making it easier to access, query, and manipulate, which will be utilized for formulating different medical tasks (in Section \ref{sec:benchmark}).


\paragraph{\textbf{Discussion.}} Similar to LCP, different medical specialties may requires different bi-level medical information. When constructing SPs of different specialties, it is recommend to conduct some minor structure adjustments to accommodate various diseases.

\subsection{Retrieval-Augmented Evaluation}
\label{sec:RAE}
In this section, we formulate a Retrieval-Augmented Evaluation (RAE) to achieve an automatic algorithm for clinical tasks, where the retrieval task can save labor-intensive human participation.
\subsubsection{Bi-level Retrieval} Corresponding to the bi-level data structure of SPs, we utilize a bi-level retrieval scheme to extract important information from SPs' medical records. First-level retrieval is to map the query to the coarse-grained categories, while second-level retrieval is to map the original query to the fine-grained items. Bi-level retrieval allows us to accurately obtain relevant information from SPs, \text{e.g.}, CT report and symptom description. 

\subsubsection{Multi-agent Environment Simulator}
Due to the complexity of clinical diagnostic tasks, we design a multi-agent framework to simulate the interactive environment between SPs and the doctor agent. Such framework supports three fundamental medical tasks, \textit{i.e.}, single-turn clinical QA, clinical diagnostic reasoning, and multi-turn diagnostic dialogue, which will be introduced in Section \ref{sec:benchmark}. The overall framework is illustrated in Figure \ref{fig:multi-agent}. 

\textbf{Bi-level Retrieval Agents.} Intent recognition aims to understand the query to achieve first-level retrieval, while query parser is for second-level retrieval. The implementation of these modules relies on the in-context capabilities of LLMs, where task-specific prompts are designed for these agents (as shown in Figure \ref{fig:prompt} in the Appendix). In particular, bi-level retrieval agents can extract important medical information from SPs as the context for clinical QA and reasoning tasks. For diagnostic dialogue, it could simulate the realistic clinical interactive environment, including conversation termination and interaction between SPs and the doctor agent.

\textbf{Retrieval-Augmented Evaluator.} As shown in Figure \ref{fig:multi-agent}, for clinical QA and reasoning tasks, both the context and ground-truth can be retrieved by bi-level retrieval agents, where RAE can achieve automatic evaluation by QA and reason scoring. For diagnostic dialogue, the doctor agent's query could reflect its clinical behaviors, where RAE can provide a comprehensive evaluation according to LCP (as shown in Table \ref{tab:system}).

\paragraph{\textbf{Discussion.}} When applying retrieval augmented evaluation to different medical specialties, both prompts for bi-level retrieval agents and RAE should be adjusted for adaptation. It is recommend to follow clinical practice pathways for different specialties. 

\subsection{RJUA-SPs Benchmark}
\label{sec:benchmark}
By leveraging the proposed paradigm including LCP, SPs, and RAE, we construct a benchmark termed RJUA-SPs, which is derived from the real patient records with doctors' clinical experience.
\subsubsection{Benchmark Construction} To ensure the data quality of our benchmark, we design a two-step data processing framework. 

\paragraph{\textbf{Step 1: Data Collection and Preprocessing}}
Clinical physicians firstly select 45 common urological diseases and then collect actual communication statements, diagnostic, and treatment information of patients afflicted with these diseases. Then, expert-level standardized transcriptions are conducted to create 45 standardized patients. The participation of clinical physicians consists of following stages: 

(1) \underline{Basic Information Statistical Modeling.} Clinical physicians need to create models reflecting the demographic and health characteristics, \textit{e.g.}, age, gender, overall health status, potential comorbidities, and medical or surgical history typically associated with patients suffering from these conditions. 

(2) \underline{Subjective Consultation Statistical Modeling.} Clinical physicians are required to construct a nuanced subjective narrative for the primary patient, involving authentic initial descriptions, communicative interactions, and intentions observed in real-world consultations. Meanwhile, such statement should incorporate crucial symptomatic expressions and authentically replicate the natural tone and reasoning of patients in clinical settings. 


(3) \underline{Objective Evidence Statistical Modeling.} Clinical physicians should synthesize and codify objective medical findings from physical assessments, laboratory tests, radiologic results, and histopathological studies of real-life cases, which can maintain the completeness of medical evidences and mirror the factual information a clinician should gather from a patient. 

(4) \underline{Clinical Pathway Statistical Modeling.} This stage involves the systematic organization and classification of SPs data.
Following rigorous clinical diagnostic protocols and routines, clinical physicians should meticulously re-organize SPs' medical record with the comprehensive continuum of care, which is from the initial diagnosis through to the therapeutic endpoint.



\paragraph{\textbf{Step 2: Completeness and Clinician Validation.}}
For each SP record, at least three rounds of manual calibration and validation are conducted. First, one urology expert performs medical logic calibration. Then, algorithm engineers carry out model tuning, followed by a second validation by the team of medical experts. This process involves verifying the clinical authenticity of the patient's information, the accuracy of specialist terminology, the rationality of question and answer exchanges, the reasoning process, and the correctness of the diagnosis.

\subsubsection{Fundamental Medical Tasks}
RJUA-SPs benchmark supports three fundamental medical tasks including single-turn medical QA, multi-turn diagnostic dialogue, and medical diagnostic reasoning. Figure \ref{fig:tasks} visualizes the examples for these tasks.
\paragraph{\textbf{Single-turn Clinical QA}} Let SPs be a structured knowledge base for a QA system. Each SP contains the complete medical information for clinical diagnosis. The goal of clinical QA is that the doctor agent should generate a answer (A) given (Q, S, T, E) as the context. Notably, the context generation can be achieved by the proposed RAE, \textit{i.e.}, retrieving medical information from SPs' records. Besides, the automatic evaluation can be conducted by the comparison between (A) and (R).

\paragraph{\textbf{Clinical Diagnostic Reasoning}} SPs can further work as a structured knowledge base for a diagnostic reasoning system. Given the initial query (Q) posed by the patient, the clinician should ask some possible symptoms (S) and can guide patients to take the effective and necessary tests (T) and examinations (E). Thus, the doctor agent should dominate a certain level of reasoning ability from a basic query (Q) to the medical information (S, T, E) awaiting verification. Here the automatic evaluation is to measure whether the doctor agent can make a request for the important medical information (S, T, E). Similarly, both the context generation and automatic evaluation can be conducted by RAE.

\paragraph{\textbf{Multi-turn Diagnostic Dialogue}} One of the most challenging medical task is multi-turn diagnostic dialogue, which is also supported by our RJUA-SPs benchmark. Let ($Q$) be the initial inquiry posed by the patient. The doctor agent should ask questions to gradually collect useful medical information (S, T, E) from the patient during the communication, and finally make the final diagnostic and treatment results. The patient is to understand the doctor’s demands and provide the relevant information, which can be simulated via a retrieval manner. Interestingly, both single-turn clinical QA and clinical diagnostic reasoning are the important sub-tasks of multi-turn diagnostic dialogue. Therefore, the dialogue simulation and automatic evaluation can be conducted using the proposed RAE, as shown in Figure \ref{fig:multi-agent} and Table \ref{tab:system}, respectively.
\subsubsection{Benchmark Characteristics} The dataset is characterized by the following three attributes. Details can refer to the Appendix.

\textbf{Specialization.} The benchmark construction is guided by expert-level clinicians in the field of urology, which have more than 10 years of clinical experiences, \textit{i.e.}, about 497,000 outpatient and 27,000 surgical procedures per year.

\textbf{Comprehensiveness}
SPs' medical records encompass 85\% (34 out of the top 40) of the most prevalent diseases of urology, which refers to about 96.8\% of real patients (309,191 out of 319,401).

\textbf{Scalability}
RJUA-SPs adopts a bi-level structure enabling dataset to scale horizontally and vertically. For example, it is easy to add new attribute information for a SP, as well as adding a new SP.

\textbf{Reliability}
At least three rounds of manual calibration and validation are conducted for each SP record, involving the clinical authenticity of the patient's information.

\section{Experiment}
In this study, we evaluate the clinical capabilities of the
latest and most advanced LLMs on the RJUA-SPs for three fundamental medical tasks. In particular, we exhibit some invalid cases for the multi-turn diagnostic dialogue task.to further explore LLMs' capabilities.

\subsection{Baselines}
The tested model include the following open-source and close-source LLMs, which are employed to simulate doctors.

\paragraph{\textbf{HuatuoGPT-II}} HuatuoGPT-II \cite{chen2023huatuogpt} is a specific LLM for medical consultation. HuatuoGPT-II leverages a single-stage domain adaption way to unify pre-training and supervised stages, which is combined with a data priority sampling strategy for adaptive adjustment of data mixture. We utilize the HuatuoGPT-II-13B version.

\paragraph{\textbf{Baichuan}} Baichuan~\citep{baichuan2023baichuan2} is an open-source large-scale multilingual language model containing 13 billion parameters, which is trained from scratch on 2.6 trillion tokens. This model excels at dialogue and context understanding. We utilize Baichuan2-7B and Baichuan2-13B versions.

\paragraph{\textbf{ChatGLM}} ChatGLM~\citep{zeng2022glm} is an open-source bilingual language model, which contains 6.2 billion parameters with specific optimization, involves supervised fine-tuning, feedback bootstrap, and reinforcement learning with human feedback. We include ChatGLM3 as a baseline for evaluations. We utilize Chatglm2-6B and Chatglm3-6B versions.

\paragraph{\textbf{Qwen}} QWen~\citep{qwen} is a comprehensive LLM series, which encompasses distinct models with varying parameter scales. The base models consistently demonstrate superior performance across a multitude of downstream tasks. We utilize the Qwen-6B version.

\paragraph{\textbf{GPT-3.5-turbo and GPT-4}} GPT series models \cite{brown2020language} are advanced language models developed by OpenAI. One of the key features of GPT series models is its ability to perform a wide range of natural language processing tasks, such as language translation, summarization, question answering, and text completion. 


\begin{table}[!t]
  \caption{Performance (\%) on the single-turn medical QA task. {\color{red} Red} denotes the best and {\color{blue} blue} denotes the second best.}
  \label{tab:qa_result}
  \vspace{-10pt}
  \begin{tabular}{l|c|cc}
    \toprule[1.5pt]
    Model & Size &Diagnosis Acc.  & Treatment Acc.\\ \hline
    HuatuoGPT-II &13B & 45.45 & 11.36\\
    Baichuan2    &chat-7B &54.55 & {\color{red}\textbf{22.73}}  \\
    Baichuan2    &chat-13B&{\color{red}\textbf{65.91}} &{\color{red}\textbf{22.73}}  \\
    Chatglm2     &6B &40.91 &9.09   \\
    Chatglm3     &6B &40.91 &{\color{blue}\textbf{18.18}}   \\
    Qwen         &chat-6B &45.45 &{\color{blue}\textbf{18.18}} \\
    GPT3.5  & Turbo  &52.27 &{\color{blue}\textbf{18.18}} \\
    GPT4         &-  &{\color{blue}\textbf{63.63}} &{\color{blue}\textbf{18.18}} \\
  \bottomrule[1.5pt]
\end{tabular}

\end{table}
\begin{table}[!t]
  \caption{Performance on the medical diagnostic reasoning task. {\color{red} Red} denotes the best and {\color{blue} blue} denotes the second best.}
  \label{tab:reason_result}
  \vspace{-10pt}
  \begin{tabular}{l|c|c}
    \toprule[1.5pt]
    Model & Size & Accuracy (\%)  \\ \hline
    HuatuoGPT-II &13B & 14.16  \\
    Baichuan2    &chat-7B  & {\color{blue}\textbf{18.62}} \\
    Baichuan2    &chat-13B & 6.12  \\
    Chatglm2     &6B  & 10.59  \\
    Chatglm3     &6B  & 10.71 \\
    Qwen         &chat-6B  & 17.60\\
    GPT3.5 & Turbo   & 11.22  \\
    GPT4 &-   & {\color{red}\textbf{20.15}}  \\
  \bottomrule[1.5pt]
\end{tabular}
\end{table}

\begin{table*}[!t]
  \caption{Performance on the multi-turn diagnostic dialogue. {\color{red} Red} denotes the best and {\color{blue} blue} denotes the second best.}
  \vspace{-10pt}
  \label{tab:dialogue}
  \begin{tabular}{c|c|c|c|c|c|c|c|c}
    \toprule[1.5pt]
    \multirow{2}{*}{Capability} & HuatuoGPT-II  & Baichuan2 & Baichuan2 & Chatglm2 & Chatglm3 & Qwen & GPT3.5 & GPT4\\ \cline{2-9}
      & 13B              & chat-7B        & chat-13B       & 6B       & 6B       &chat-6B & Turbo & -\\ \hline
    Information Completeness   &3.20 & 5.31&2.33 & 6.33& {\color{blue}\textbf{10.92}}&4.08 & 7.13 & {\color{red}\textbf{15.13}}\\\hline
    Behavior Standardization        
    & \XSolidBrush& \XSolidBrush& \XSolidBrush& \XSolidBrush& \XSolidBrush& \XSolidBrush &\XSolidBrush & \XSolidBrush\\\hline
    Guidance Rationality    &3.21  &8.21&2.14 &4.29 &5.71 &3.93& {\color{blue}\textbf{13.21}} &{\color{red}\textbf{25.36}}\\\hline
    Diagnostic Logicality                 &25.00 &18.18&20.45&22.72 &20.45 & {\color{blue}\textbf{27.27}}& 22.73 & {\color{red}\textbf{50.00}}\\\hline
    Treatment Logicality               &4.54  &4.54 &2.27 &0 &9.10 &4.55&{\color{red}\textbf{6.82}} &{\color{red}\textbf{6.82}}\\\hline
    \makecell{Clinical Applicability \\Avg. Round (8.39)  }         &1.95 &2.34&2.20 &4.41 & {\color{red}\textbf{6.97}}&3.14 &2.07 & {\color{blue}\textbf{4.00}}\\
  \bottomrule[1.5pt]
\end{tabular}
\end{table*}

\subsection{RAE-based Automated Metrics}
\noindent \textbf{Single-turn Clinical QA.} We calculate the diagnosis and treatment accuracy, where RAE can retrieve the ground-truth from SPs.

\noindent \textbf{Medical Diagnostic Reasoning.} We calculate the accuracy of the recommended laboratory test and examination, where RAE can retrieve the ground-truth from SPs.

\noindent \textbf{Multi-turn Diagnostic Dialogue.} Importantly, RAE can measure the clinical capabilities of LLMs by the retrieval task. The detailed metrics are shown in Table \ref{tab:system}.

\subsection{Main Results and Discussions}
Table \ref{tab:qa_result}, \ref{tab:reason_result}, \ref{tab:dialogue} shows the results of automatic evaluation on single-turn medical QA, medical diagnostic reasoning, and multi-turn diagnostic dialogue, respectively. Overall, GPT4 performs the best performance for these medical tasks, but it still exhibits some weakness points. Now, we report the clinical capabilities through the observations from these tasks.

\textit{\textbf{LLMs struggle with the clinical specialty.}}  As shown in Table \ref{tab:qa_result}, Baichuan2-13B performs best on the QA task for both diagnosis and treatment, while the top-performing model GPT-4 achieves the second best. Although providing the complete medical information of the patients, all models show a lower diagnosis accuracy below 70\% and a extremely lower treatment accuracy below 25\%. Notably, the medical-specific open-source model, HuatuoGPT-II-13B, and the close-source model, GPT-3.5-turbo, achieve only 45.45\% and 52.27\% diagnosis accuracy, as well as 11.36\% and 18.18\% treatment accuracy. There is a significant performance gap between the diagnosis and treatment, indicating that practical clinical capabilities of LLMs in real-world medical scenarios remains significantly limited. 

\textit{\textbf{LLMs are weak in collecting medical information of patients.}} As shown in Table \ref{tab:reason_result}, most models exhibit a lower reasoning accuracy below 20\%. Only GPT4 can obtains the best (20.15\%). When providing the initial query and symptoms of the patients, these models cannot recommend effective advice for medical tests and examinations. Therefore, they may fail to collect sufficient objective information as medical evidences during clinical applications.

\textit{\textbf{LLMs are weak in multi-turn diagnostic dialogue capabilities.}} As shown in Table \ref{tab:dialogue}, GPT4 exhibits the best diagnostic dialogue capabilities. In detail, all models cannot collect complete medical information of patients via consultations, \textit{i.e.}, lower information completeness below 16\%. Besides, these models also obtain lower guidance rationality, indicating that they may fail to acquire appropriate test or examination reports of patients, which is also indicated by Table \ref{tab:reason_result}. The average dialog rounds of these models are greatly less than the average dialog round of clinicians (8.39), as explained in Section \ref{sec:case}. Chatglm3 conducts more consultation rounds because it poses many ineffective questions.

\textit{\textbf{LLMs fail to follow the order of collecting medical information.}} As shown in Table \ref{tab:dialogue}, clinical behavior standardization of all models (including GPT4) exhibit inconsistency with the ground-truths from clinicians. These LLMs tend to require a random test (or examination) report from a patient.

\subsection{Case Study}
\label{sec:case}
In this section, we showcase some examples of invalid dialogue content, mainly including the following cases. All results are shown in Appendix due to limited space.
\begin{itemize}[leftmargin=*]
    \item The doctor agent often repeatedly asks the same question during the clinical diagnostic process, as shown in Table \ref{tab:repeat}.
    \item Given a test report, the doctor agent tends to immediately produce the final diagnosis results, while ignoring collecting more medical information of patients, as shown in Table \ref{tab:without}.
    \item The doctor agent cannot acquire the specific test when collecting medical information of patients, as shown in Table \ref{tab:name}.
\end{itemize}

\subsection{Impacts of Guidance Rationlity}
\begin{figure}[!t]
  \centering
  \includegraphics[width=1\linewidth]{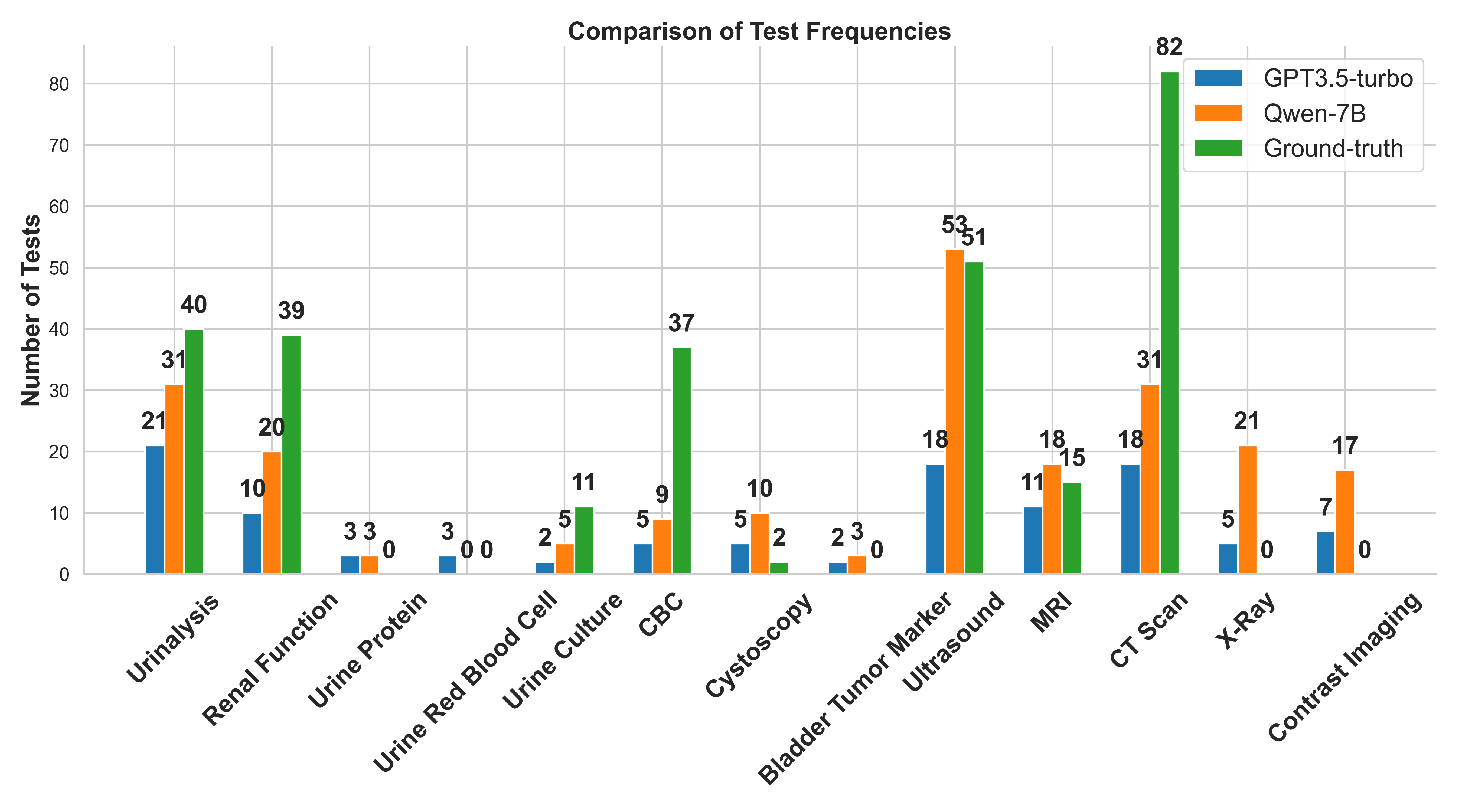}
  \caption{Comparisons of Medical Test Guidance.}
  \label{fig:test_guidance}
\end{figure}

In this part, for different LLMs, we visualize the recommended frequencies of the commonly used medical tests.
As shown in Figure \ref{fig:test_guidance}, the comparison shows that both LLMs have discrepancies when compared to the ground-truth. 
LLMs report some tests while the ground-truth reports none, indicating unnecessary medical tests recommend by LLMs. For example, 
both LLMs suggest the Urine Protein test, but the ground-truth indicates that this test was not important (number is 0). This could mean that the LLMs may recommend unnecessary tests in clinical case. The ground-truth shows a significantly higher frequency than each LLM, with GPT3.5-turbo showing the largest gap.

\begin{figure}[!t]
  \centering
  \includegraphics[width=1\linewidth]{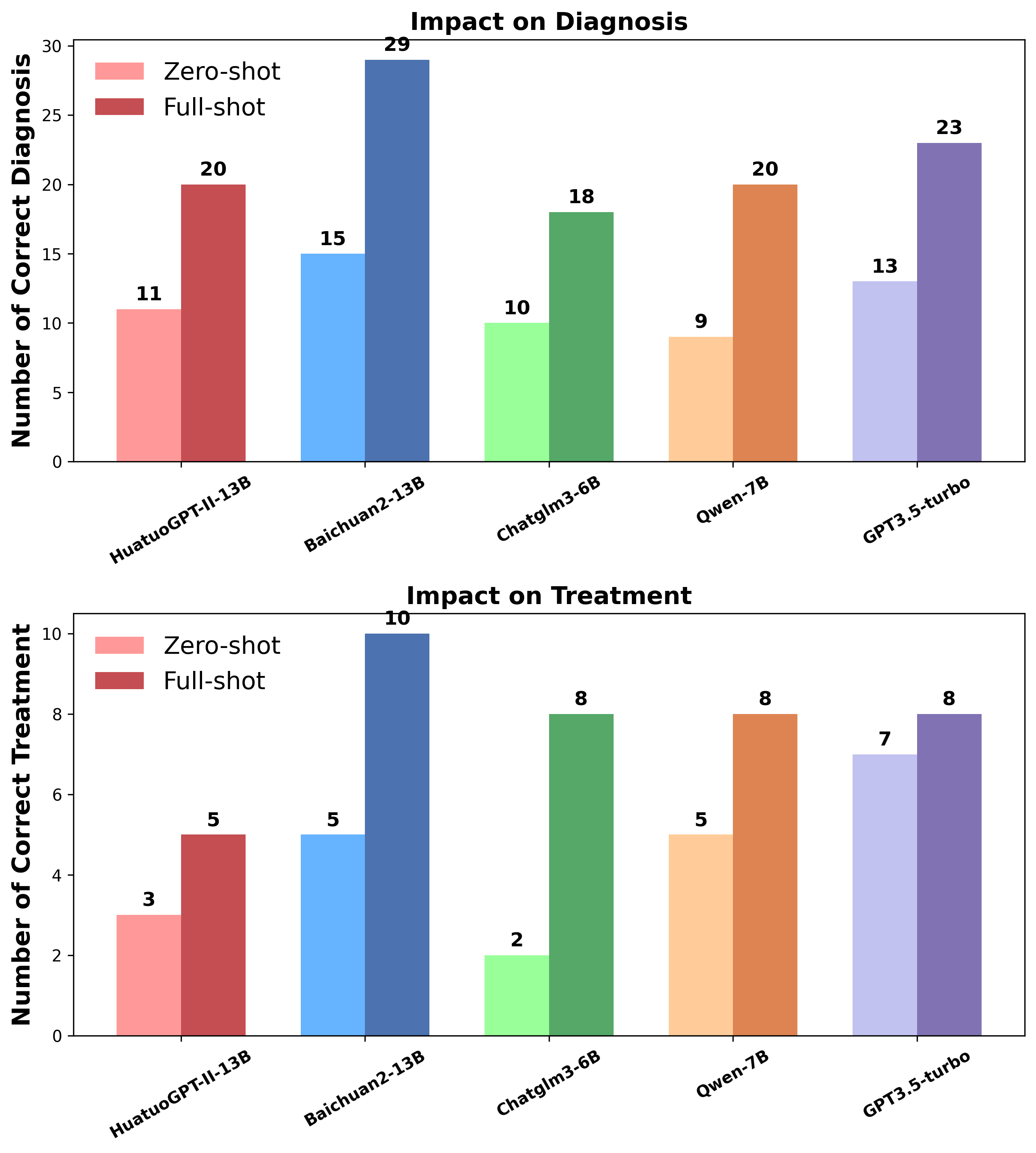}
  \vspace{-10pt}
  \caption{Impacts of Zero-shot and Full-shot settings.}
  \label{fig:zero}
\end{figure}

\subsection{Impact of Information Completeness}
In this section, we investigate the importance of information completeness of patients during clinical scenarios. We consider two settings: (1) Zero-shot setting where the doctor agent only knows the initial query of patients without any other medical information. (2) Full-shot setting where the doctor agent knows all medical information of patients. As shown in Figure \ref{fig:zero},
all models perform poorly on the zero-setting with
with a moderate amount of medical information. After providing more medical information, the accuracy of diagnosis and treatment can be improved, indicating the importance of information completeness.

\section{Conclusion}
In this work, we propose an automatic evaluation paradigm tailored to assess the LLMs' capabilities in delivering clinical services. The evaluation paradigm contains three basic elements: metric, data, and algorithm. Firstly, we formulate a LLM-specific clinical pathway based on professional clinical practice pathways. Then, Standardized Patients from the medical education are introduced to collect medical data, ensuring the completeness of the evaluation procedure. Leveraging these steps, we develop a multi-agent framework to simulate the interactive environment between SPs and a doctor agent, where a Retrieval-Augmented Evaluation is designed to determine whether the behaviors of a doctor agent are in accordance with LCP. Applying such paradigm, we construct an RJUA-SPs benchmark in the field of urology. Extensive experiments can provide more insights for LLMs' safe and reliable deployments in clinical applications.



\bibliographystyle{ACM-Reference-Format}
\bibliography{sample-base}

\clearpage
\appendix
\section{Appendix}
\subsection{Benchmark Characteristics}

The dataset is derived from the real patient diagnosis and treatment process in doctors' reference clinical experience. Drawing inspiration from the Standardized Patient (SP) model commonly employed in clinical education, we have constructed a virtual patient clinical database. The dataset has been developed in collaboration with the expert team of urologists from Renji Hospital in Shanghai, ensuring that no private medical or patient data is involved.
The dataset is characterized by the following three attributes:
\paragraph{\textbf{Specialization}} 
The benchmark construction is guided by expert-level clinicians in the field of urology. which can ensure the data quality of SPs with medical professionalism. These clinicians can conduct about 497,000 outpatient and 27,000 surgical procedures per year, thus accumulating a wealth of clinical experience (\textit{i.e.}, over 10 years).
\paragraph{\textbf{Comprehensiveness}} 
The standardized patients constructed in this dataset encompass 85\% (34 out of the top 40) of the most prevalent diseases among patients visiting the Urology Department of Renji Hospital in Shanghai. In terms of patient numbers, the coverage is even more impressive, reaching 96.8\% (309,191 out of 319,401). The clinical reference data for the standardized patients, which includes imitations of real patient discourse, outpatient and emergency diagnostic and treatment information, as well as all objective materials that could be provided by hospitalized surgical patients, bears high authenticity, practical significance, and application value.
\paragraph{\textbf{Scalability}} 
RJUA-SPs adopts a bi-level structure enabling dataset to scale horizontally and vertically. For example, it is easy to add new attribute information for a SP, as well as adding a new SP.

\paragraph{\textbf{Reliability}}
For each standardized patient data, at least three rounds of manual calibration and validation are conducted. First, the Renji urology expert team performs medical logic calibration. Then, algorithm engineers carry out model tuning, followed by a second validation by the medical team. This process involves verifying the clinical authenticity of the patient's information, the accuracy of specialist terminology, the rationality of question and answer exchanges, the reasoning process, and the correctness of the diagnosis.

\subsection{Prompts for Multi-Agent Framework}
\begin{figure}[!t]
  \centering
  \includegraphics[width=0.9\linewidth]{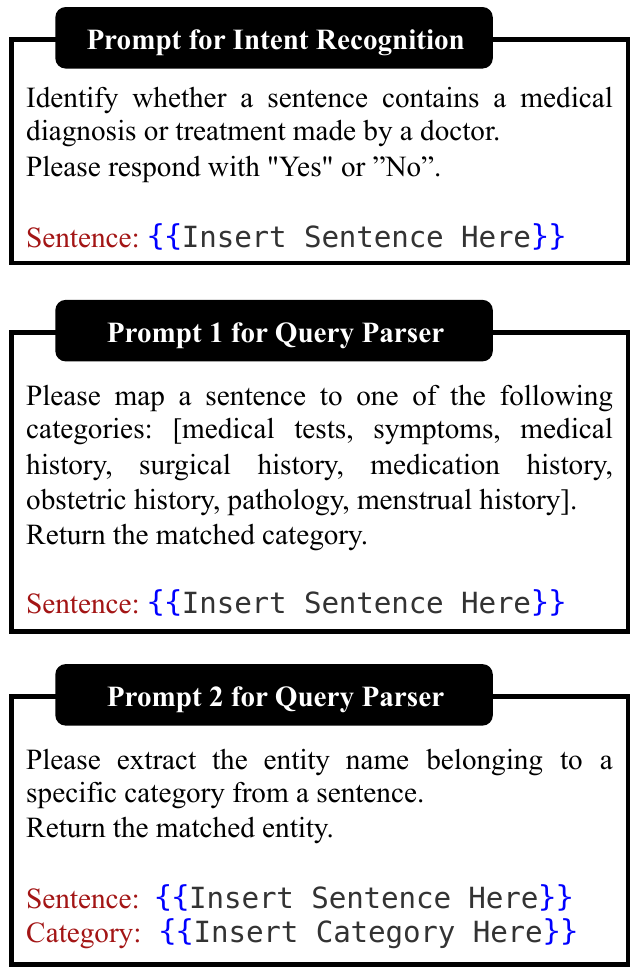}
  \caption{Prompt templates for the multi-agent LLMs. Elements in double braces $\{\{\}\}$ are replaced with specific contents.}
  \label{fig:prompt}
\end{figure}

We design some prompt templates for the retrieval-augmented evaluation, as shown in Figure \ref{fig:prompt}. This allowed us to achieve automatic evaluation via the retrieval task.

\subsection{Clinical Pathway for Data Collection}
\begin{figure*}[!t]
  \centering
  \includegraphics[width=1\linewidth]{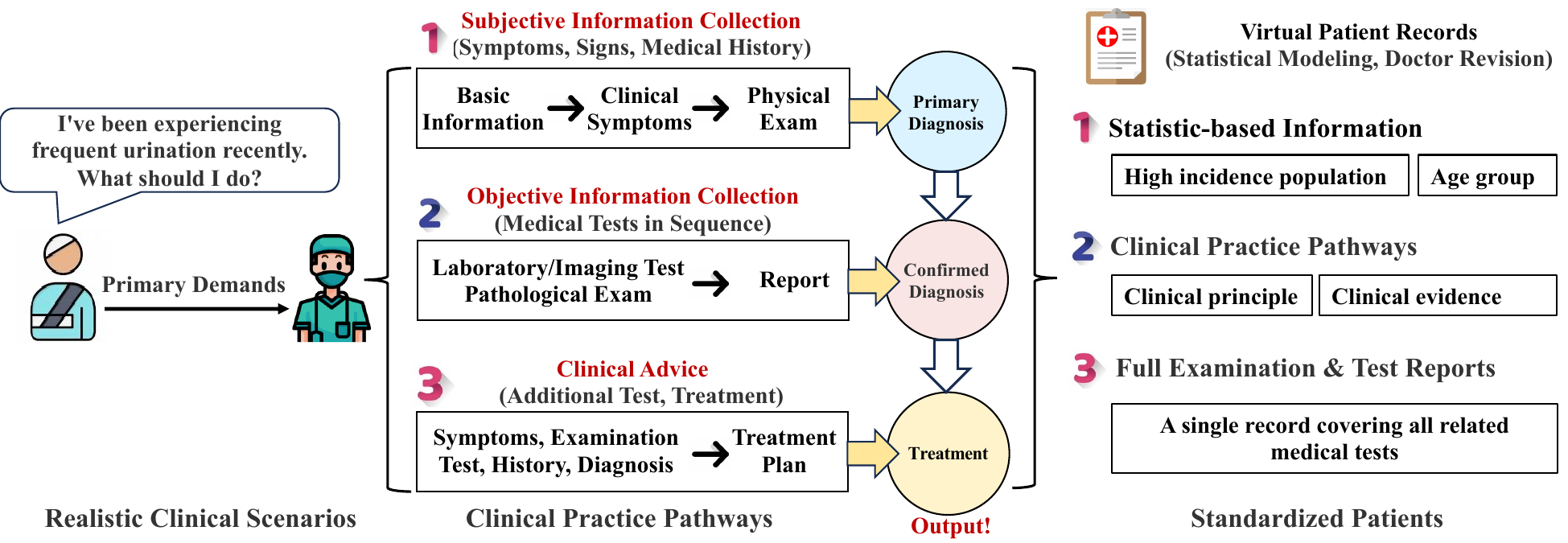}
  \caption{The overall paradigm. Following clinical practice pathways in realistic scenarios, medical experts need to re-organize clinical records to formulate SPs, which are equipped with virtual yet complete medical records.}
  \label{fig:cpp}
\end{figure*}

In this section, we visualize a detailed pipeline for collecting SPs data from the realistic clinical scenarios. The overall pipeline is shown in Figure \ref{fig:cpp}.

\subsection{Details of RAE}
\begin{figure*}[!t]
  \centering
  \includegraphics[width=0.9\linewidth]{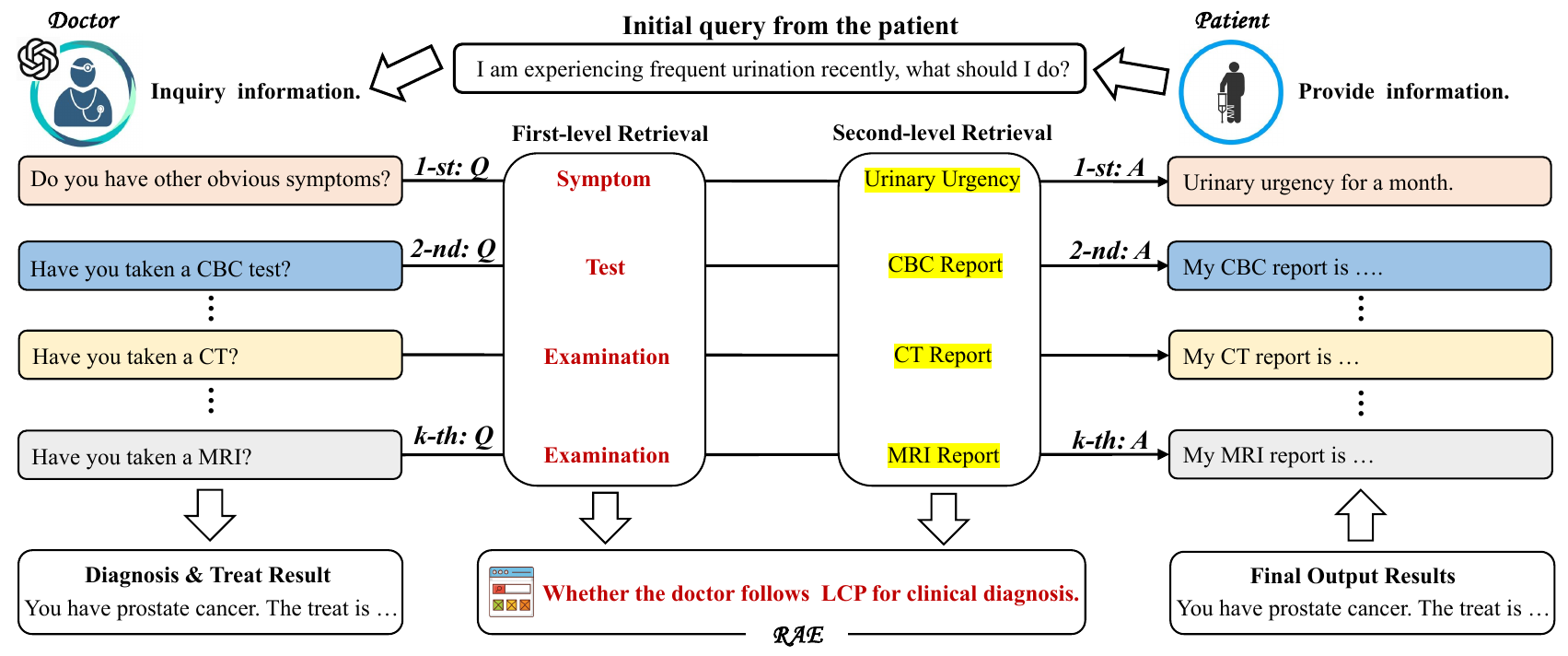}
  \caption{RAE Example for multi-turn diagnostic dialogue task. The doctor agent should query medical information from the patient. The patient provide the response by RAE retrieving from SPs' recors. RAE can measure whether the doctor follows LCP to conduct clinical behaviors.}
  \label{fig:RAE}
\end{figure*}

In this section, we illustrate the detailed steps of RAE for multi-turn diagnostic dialogue. The simulated dialogue procedure is shown in Figure \ref{fig:RAE}.

\subsection{A full Example of SP}
In this section, we provide a full version of a SP medical record, as shown in Table \ref{tab:sps}.

\begin{table*}[!t]
  \centering
  \caption{An example of the medical record of a virtual SP. We maintain the information completeness for the SPs, thus SPs can would support a full-process diagnostic procedure. Please note that some information is omitted or reduced due to the limited space, such as the report result. The full medical record can be found in the released benchmark.}
    \begin{tabular}{c|c|p{20.25em}|c|c}
    \hline
    \textbf{1-st Category} & \textbf{2-nd Category} & \textbf{Report Result}    &\textbf{Constraints} & \textbf{Importance} \\
    \hline
    Initial Query &       & After my mom peed, she noticed that the toilet felt like there was blood in it.
     &       & -  \\
    \hline
    Patient's Intent &       & Prefer conservative treatment &       &       \\
    \hline
    Expected Rounds &       & 7 Rounds    &       &        \\
    \hline
    Age &       & Female     &       &     \\
    \hline
    Gender &       & 84 Years Old   &       &      \\
    \hline
    Diagnosis & Core Diagnosis & Bladder Cancer&       &  $\star \star \star$  \\
    \hline
    \multicolumn{5}{c}{\color{red}\textbf{Key Elements and Correct Order of Questions for Confirming the Diagnosis}}\\
    \hline
    Symptom & Gross Hematuria & Consistent blood in urine, color is pink. & 12 Days & $\star \star$  \\
    \hline
    Symptom & Urgency & None     &       & $\star \star$ \\
    \hline
    Symptom & Fever   & None     &       & $\star \star$  \\
    \hline
    Symptom & Difficulty Urinating & None     &       & $\star \star$   \\
    \hline
    Test & Urinalysis & Urine color is brownish red, specific gravity 1.016, pH 6.0. & 7 Days Ago & $\star \star$ \\
    \hline
    Test & Urinary NMP22 & Positive    & 7 Days Ago & $\star \star$  \\
    \hline
    Examination & Urinary Ultrasound & Normal size and shape, clear capsule contour. & 7 Days Ago & $\star \star$  \\
    \hline
    Examination & Bladder Enhanced MRI & Multiple occupying lesions on the left side. & 7 Days Ago & $\star \star$  \\
    \hline
    Examination & CT Urography & Posterior wall of the bladder. Left kidney cyst. & 7 Days Ago & $\star \star$  \\
    \hline
    Correct Order &       & Urinary Ultrasound > Enhanced MRI = CT Urography &       & $\star \star$  \\
    \hline
    Treatment &       & TURBT &       & $\star \star \star$  \\
    \hline
    \multicolumn{5}{c}{\color{red}\textbf{Key Elements of Treatment Plan}}    \\
    \hline
    Surgical History & None &  &       & $\star \star$  \\
    \hline
    Medication History & None &  &       & $\star \star$  \\
    \hline
    General Condition &       & Sleep is fairly good, no unusual diet, no significant weight change. &       &  $\star $  \\
    \hline
    Medical History & Hypertension & Blood pressure controlled around 140/90 &       & $\star $  \\
    \hline
    Medical History & Diabetes Mellitus & Negative    &       & $\star $   \\
    \hline
    Medical History & Coronary Heart Disease & Negative    &       & $\star $  \\
    \hline
    Marital \& Childbearing & Married with Child & \multicolumn{1}{c|}{} &       &         \\
    \hline
    Menstrual History & Menopause & \multicolumn{1}{c|}{} &       &          \\
    \hline
    Preoperative Test  & Complete Blood Count   & White blood cell count 4.69x10\^9/L. & 1 Day Ago   & $\star $    \\
     \hline
    Preoperative Test  & Liver Function   & Total bilirubin 12.9 $\mu$mol/L, direct bilirubin 3.2 $\mu$mol/L.& 1 Day Ago   & $\star$   \\
    \hline
    Preoperative Test  & Renal Function   & Urea 5.60mmol/L, creatinine 48.0 $\mu$mol/L. & 1 Day Ago   & $\star$ \\ 
    \hline
    Preoperative Test  & Electrolytes   & Potassium 3.89mmol/L, sodium 141.40mmol/L, chloride 105.80mmol/L; & 1 Day Ago   & $\star $ \\
    \hline
    Preoperative Test  & Fasting Blood Glucose  & Fasting blood glucose 6.56mmol/L. & 1 Day Ago   & $\star $     \\
    \hline
    Preoperative Test  & Coagulation Function  & Prothrombin time 11.80 seconds. & 1 Day Ago   & $\star $   \\
    \hline
    Preoperative Test  & B-type Natriuretic Peptide & B-type natriuretic peptide 70.0pg/mL; & 1 Day Ago    \\
    \hline
      Preoperative Test & Cardiac Infarction Markers & Troponin I 0.01ng/ml, Myoglobin 15.80ng/ml. & 1 day ago & $\star $  \\
    \hline
    Preoperative Test & Hepatitis B & Hepatitis B surface antigen 0.45COI. & 1 day ago & $\star $  \\
    \hline
    Preoperative Test & Hepatitis C & Hepatitis C antibody (C) 0.04S/CO, & 1 day ago & $\star $   \\
    \hline
    Preoperative Test & HIV & HIV Ag/Ab 0.05COI & 1 day ago & $\star $  \\
    \hline
    Preoperative Test & Syphilis & Confirmatory test for syphilis negative & 1 day ago & $\star $ \\
    \hline
    Preoperative Test & Chest CT Scan & Scattered tiny nodules in both lungs. & 1 day ago & $\star $  \\
    \hline
    Preoperative Test & Echocardiography & No obvious abnormalities. & 1 day ago & $\star $  \\
    \hline
    Preoperative Test & Electrocardiogram & Sinus rhythm, low flat T waves. & 1 day ago & $\star $ \\
     \hline
    \end{tabular}%
    \label{tab:sps}%
\end{table*}

\begin{table*}
\centering
\caption{An example of utilizing Chatglm3-6B to act as a doctor. The main issue is repeatedly acquiring the same question.}
\vspace{-10pt}
\begin{tabular}{p{1.5cm}|p{13cm}}
\toprule[1.5pt]
\textbf{Patient:} & Hello, Doctor. My father has been suffering from back pain for a long time. An ultrasound performed half a month ago revealed a tumor on his left kidney. \\
\hline
\textbf{Doctor:} & \textcolor{red}{Are there any symptoms of back pain?} \\
\hline
\textbf{Patient:} & The pain and soreness in the left side of the back have significantly worsened. \\
\hline
\textbf{Doctor:} & Have you undergone any related examinations? \\
\hline
\textbf{Patient:} & I had a renal enhancement CT scan.  The result is ... \\
\hline
\textbf{Doctor:} & \textcolor{red}{Are there any symptoms of back pain?} \\
\hline
\textbf{Patient:} & The pain and soreness in the left side of the back have significantly worsened. \\
\hline
\textbf{Doctor:} & Have you undergone any related examinations? \\
\hline
\textbf{Patient:} & Tumor marker tests.  The result is ... \\
\hline
\textbf{Doctor:} & \textcolor{red}{Are there any symptoms of back pain?} \\
\hline
\textbf{Patient:} & The pain and soreness in the left side of the back have significantly worsened. \\
\hline
\textbf{Doctor:} & Have you undergone any related examinations? \\
\hline
\textbf{Patient:} & Urinalysis. The result is ... \\
\bottomrule[1.5pt]
\end{tabular}
\label{tab:repeat}
\end{table*}

\begin{table*}
\centering
\caption{An example of utilizing GPT3.5-turbo to act as a doctor. The main issue is providing diagnosis results without acquiring more medical information of patients.}
\vspace{-10pt}
\begin{tabular}{p{1.5cm}|p{13cm}}
\toprule[1.5pt]
\textbf{Patient:} & My right side of the waist is sore, I used to have frequent kidney stone pain attacks.\\
\hline
\textbf{Doctor:} & Have you had an ultrasound examination?\\
\hline
\textbf{Patient:} & I had an ultrasound done 10 days ago. The result is ...\\
\hline
\textbf{Doctor:} & According to your examination results, you have stones in your right kidney and an enlarged prostate. I suggest you drink more water and consider taking medication to dissolve the stones.\\
\bottomrule[1.5pt]
\end{tabular}
\label{tab:without}
\end{table*}

\begin{table*}
\centering
\caption{An example of utilizing Chatglm2-6B to act as a doctor. The main issue is that the doctor cannot provide a specific test name to collect information of patients.}
\vspace{-10pt}
\begin{tabular}{p{1.5cm}|p{13cm}}
\toprule[1.5pt]
\textbf{Patient:} & Hello, Doctor. I've had a dull pain in my right flank for several months.\\
\hline
\textbf{Doctor:} & Have you had any particular tests done? \textcolor{red}{Without specific test names?} \\
\hline
\textbf{Patient:} & I had an ultrasound of the urinary system one month ago. The result is ...\\
\hline
\textbf{Doctor:} & May I ask if you have undergone any recent examinations or tests? \textcolor{red}{Without specific test names?}\\
\bottomrule[1.5pt]
\end{tabular}
\label{tab:name}
\end{table*}

\clearpage
\begin{table*}[!t]
\centering
\caption{An example of Baichuan2-13B-chat for the single-turn medical task on RJUA-SPs.}
\label{qa_instance_baichuan2_13}
\vspace{-10pt}
\begin{tabular}{p{3cm}|p{10cm}}
\toprule[1.5pt]
\textbf{Question + Context} & Hello doctor, here are the symptoms: gross hematuria without back pain, fever, or dysuria.
The following examinations were conducted:
Urinary system ultrasound 7 days ago: multiple occupancies in the bladder, please correlate clinically.
Bladder MRI enhancement 7 days ago: several occupancies on the left side and posterior wall of the bladder, bladder cancer is considered likely, please correlate with cystoscopy. Uterine fibroid. Abnormal signal shadow near the left femur.
CT urography (CTU) 7 days ago: several occupancies on the left side and posterior wall of the bladder, bladder cancer is considered likely, please correlate with cystoscopy. Routine urine test 7 days ago: urine color brownish-red, urine turbidity turbid, specific gravity 1.016, urine pH 6.0, urine nitrite neg, urine leukocyte esterase neg, urine protein 1+1, urine glucose neg, urine ketones neg, urobilinogen norm, urine bilirubin neg, urine occult blood 3+t, microscopic examination red blood cells >100.0/HPt, microscopic examination white blood cells not seen. What disease do I have and how should it be treated?\\
\midrule
\textbf{Answer} & 
    Based on the patient's symptoms and examination results, the most likely disease is bladder cancer. The best treatment plan is surgical treatment, such as cystectomy, and may need to be combined with chemotherapy to eliminate possible residual cancer cells. \\ \midrule
\textbf{Ground-truth} & 
    Malignant tumor of the bladder, treated with transurethral resection of bladder tumor (TURBT) \\
\bottomrule[1.5pt]
\end{tabular}
\end{table*}

\begin{table*}[!t]
\centering
\caption{An example of Baichuan2-13B-chat for the medical reason task on RJUA-SPs.}
\label{reason_instance_baichuan2_13}
\vspace{-10pt}
\begin{tabular}{p{3cm}|p{10cm}}
\toprule[1.5pt]
\textbf{Question + Context} & Hello doctor, 12 days ago I noticed that the toilet was red after my mother urinated. There was gross hematuria throughout, with a pinkish color, and the color was deeper upon waking up in the morning. What tests or examnibations dose she should take?\\
\midrule
\textbf{Recommendations} & Based on your mother's symptoms, I suggest that you undergo the following necessary medical examinations and tests:
1. Urine dipstick test
2. Urine sediment microscopy
3. Complete blood count (CBC)
4. Kidney function tests (serum creatinine and blood urea nitrogen, BUN)
5. Enhanced CT scan of the upper abdomen (abdominal computed tomography, CT scan with contrast)
     \\ \midrule
\textbf{Ground-truth} & 
    Routine urinalysis, urinary system ultrasound, enhanced MRI of the bladder.\\
\bottomrule[1.5pt]
\end{tabular}
\end{table*}

\begin{table*}[!t]
\centering
\caption{An example of Baichuan2-13B-chat for the medical reason task on RJUA-SPs.}
\label{dialogue_instance_baichuan2_13}
\vspace{-10pt}
\begin{tabular}{p{3cm}|p{10cm}}
\toprule[1.5pt]
\textbf{Patient:} & Hello, Doctor. 12 days ago, I noticed that the toilet was red after my mother urinated.\\
\midrule
\textbf{Doctor:} & Does your mother experience any symptoms of dysuria, increased urinary frequency, or urgency?\\ \midrule
\textbf{Patient:} & 
    There's been gross hematuria for 12 days, continuous throughout the day, with the color being pinkish. It gets darker in color in the morning.\\\midrule
\textbf{Doctor:} &  Based on your description, your symptoms could be due to a disorder of the urinary system, such as nephritis, kidney stones, or a urinary tract infection. However, this is only a preliminary assessment, and further examination is required to confirm this. I suggest that you go to the hospital as soon as possible for routine urine tests, urine cultures, kidney function tests, and an ultrasound, among other relevant examinations. This will allow doctors to make an accurate diagnosis and provide a treatment plan. In the meantime, make sure to drink plenty of water, avoid strenuous exercise and maintain good lifestyle habits.\\
\bottomrule[1.5pt]
\end{tabular}
\end{table*}

\end{document}